\documentclass[11pt, a4paper, logo, nonumbering]{imp}
\usepackage[numbers, sort&compress, square]{natbib}
\usepackage{dblfloatfix}
\usepackage{ulem}
\usepackage{caption}
\usepackage{subcaption}
\usepackage{wrapfig}
\usepackage{dramatist}
\usepackage{xspace}
\usepackage{pifont}
\usepackage{multirow}
\usepackage{makecell}
\usepackage{tcolorbox}
\usepackage{xltabular}
\usepackage{longtable}
\usepackage{pdfpages}
\usepackage{xcolor}

\definecolor{linkcolor}{rgb}{0.956,0.298,0.23}
\definecolor{urlcolor}{RGB}{122,77,182} 
\definecolor{mypurple}{RGB}{239,234,246}
\definecolor{mygray}{gray}{0.4}
\definecolor{citecolor}{HTML}{5A2BA3} 
\usepackage{tabularx}
\usepackage{hyperref}
\hypersetup{
    colorlinks=true,    
    linkcolor=citecolor,     
    filecolor=magenta,  
    urlcolor=urlcolor,      
    citecolor=citecolor,    
}

\usepackage{amsfonts}
\usepackage{amsmath}
\usepackage{amssymb}
\usepackage{lineno}

\usepackage[bottom]{footmisc}

\usepackage{CJKutf8}
\usepackage{setspace}

\usepackage{xspace}

\makeatletter
\def\@BTrule[#1]{%
  \ifx\longtable\undefined
    \let\@BTswitch\@BTnormal
  \else\ifx\hline\LT@hline
    \nobreak
    \let\@BTswitch\@BLTrule
  \else
     \let\@BTswitch\@BTnormal
  \fi\fi
  \global\@thisrulewidth=#1\relax
  \ifnum\@thisruleclass=\tw@\vskip\@aboverulesep\else
  \ifnum\@lastruleclass=\z@\vskip\@aboverulesep\else
  \ifnum\@lastruleclass=\@ne\vskip\doublerulAesep\fi\fi\fi
  \@BTswitch}
\makeatother

\addto\extrasenglish{
}

 {\begin{list}{}%
         {\setlength{\leftmargin}{#1}}%
         \item[]%
 }
 {\end{list}}
 
\reportnumber{001} 

\renewcommand{\today}{}

\title{\centering Imp: Highly Capable Large Multimodal Models \\for Mobile Devices}

\author[*]{
\small

\mbox{Zhenwei Shao},~
\mbox{Zhou Yu}$^{\dag}$,~
\mbox{Jun Yu},~
\mbox{Xuecheng Ouyang},~
\mbox{Lihao Zheng},~
\mbox{Zhenbiao Gai},
\mbox{Mingyang Wang},~
\mbox{Jiajun Ding}
\\
\small
$^1$Hangzhou Dianzi University \\
\small
\vspace{5pt}
\small
\url{https://imp-vl.github.io}
}
\correspondingauthor{\footnotesize\, $^\dag$ Project lead \& Corresponding author: Zhou Yu (yuz@hdu.edu.cn)}

\begin{abstract}
By harnessing the capabilities of large language models (LLMs), recent large multimodal models (LMMs) have shown remarkable versatility in open-world multimodal understanding. Nevertheless, they are usually parameter-heavy and computation-intensive, thus hindering their applicability in resource-constrained scenarios. To this end, several lightweight LMMs have been proposed successively to maximize the capabilities under constrained scale (e.g., 3B). Despite the encouraging results achieved by these methods, most of them only focus on one or two aspects of the design space, and the key design choices that influence model capability have not yet been thoroughly investigated. In this paper, we conduct a systematic study for lightweight LMMs from the aspects of model architecture, training strategy, and training data. Based on our findings, we obtain Imp---a family of highly capable LMMs at the 2B$\sim$4B scales. Notably, our Imp-3B model steadily outperforms all the existing lightweight LMMs of similar size, and even surpasses the state-of-the-art LMMs at the 13B scale. With low-bit quantization and resolution reduction techniques, our Imp model can be deployed on a Qualcomm Snapdragon 8Gen3 mobile chip with a high inference speed of about 13 tokens/s.   
\end{abstract}

\begin{document}
\begin{CJK*}{UTF8}{gbsn}
\maketitle

\section{Introduction}

\begin{center}
	\textit{``A very small man can cast a very large shadow.''}\footnote{This description is about a dwarf character named Tyrion Lannister in the fiction, whose nickname is \emph{Imp}.}
\end{center}
\vspace{-20pt}
\begin{flushright}
	\textit{------ George R. R. Martin, "A Clash of Kings"}
\end{flushright}

The revolution of large language models (LLMs) has greatly changed the landscape of artificial intelligence in recent years \cite {gpt2,gpt3,palm,llama1,mistral}. The latest LLMs like GPT-4 \cite{gpt4} and Gemini-1.5 \cite {gemini} exhibit remarkable versatilities and capabilities across a variety of domains and tasks. Meanwhile, there has been an increasing interest in expanding the language-only LLMs to large multimodal models (LMMs), which aims to jointly handle more modalities beyond language, such as vision \cite{gpt4v,minigpt4,qwen-vl,cogvlm,videochat}, speech \cite{pengi,speechgpt}, and their combinations \cite{videollama,macawllm,languagebind,imagebind,pandagpt}. These LMMs significantly extend the capabilities of language-only LLMs, enabling more natural and flexible interactions to accomplish complex real-world tasks.

Despite the remarkable progress achieved by these LMMs, they are usually parameter-heavy and computation-intensive in both training and inference stages, which hinder the participation of academic researchers with limited resources and pose challenges for applications in resource-constrained environments like PCs and mobile devices. In light of the limitations above, building lightweight yet strong LMMs has rapidly drawn attention from both academia and industry. Built upon the lightweight LLMs like Phi-2 (2.7B) \cite {phi2} and MiniCPM \cite {minicpm}, some representative LMMs around 3B scale have achieved comparative results to the state-of-the-art counterparts at 7B scale \cite {qwen-vl,llava1.5} by introducing carefully-designed training strategies \cite {tinygptv,varytoy}, advanced architectures \cite {mobilevlm}, and enriched training data \cite {minicpm,bunny}, respectively. 

Despite the prominence of these lightweight LMMs, most of them only focus on one or two aspects of the whole design space. It still remains unclear what the key design choices are that influence the capabilities of a lightweight LMM. Although some systematic studies have been made for the 7B/13B LMMs \cite {llava1.5,prismaticvlm}, there is no clear evidence that these empirical design choices can be directly transferred to the LMMs of much smaller sizes. To this end, we conduct a thorough study to investigate the impact of design choices of the lightweight LMMs in controlled settings. Originating from the commonly-used LLaVA-1.5 \cite {llava1.5} model, we construct a comprehensive roadmap by evaluating the impacts of different model architecture, training strategy, and training data progressively. Building upon the open-source lightweight LLMs like Qwen-1.5 \cite{qwen}, Phi-2 \cite{phi2}, and Phi-3 \cite{phi3}, we obtained a family of highly capable yet lightweight LMMs named {Imp-2B}, {Imp-3B} and {Imp-4B}, respectively. Notably, Our Imp-3B model significantly outperforms the counterparts of similar size on a broad range of LMM benchmarks and also steadily surpasses the state-of-the-art LMMs at the 13B scale. Moreover, we do not use any proprietary pretrained models or private training data to ensure reproducibility. With low-bit quantization and reduced image resolution techniques, the optimized Imp-3B model can run efficiently at about 13 tokens per second on a mobile phone equipped with a Qualcomm Snapdragon chip. The code and pretrained models are publicly available at Github\footnote{Code: \url{https://github.com/MILVLG/imp}} and  HuggingFace\footnote{Models: \url{https://huggingface.co/MILVLG}}, respectively. We hope that our work may serve as a solid baseline for future research in lightweight LMMs.

\section{Related Work}

\paragraph{Large language models (LLMs).} Due to the remarkable versatilities and  capabilities across various language tasks, LLMs have revolutionized the field of artificial intelligence in the past few years. By scaling the parameter sizes of Transformers \cite {transformer} to tens or even hundreds of billions, LLMs have exhibited emergent properties that have not been witnessed in previous {small} pretrained language models \cite {gpt1,bert,t5}. The prominent success of LLMs is first brought by commercial models like the GPT series \cite {gpt3,ouyang2022training,gpt4} and ChatGPT \cite {chatgpt}, and further accelerated by the open-sourced models such as LLaMA \cite {llama1,llama2}, Mistral \cite {mistral}, Qwen \cite {qwen}, and Baichuan \cite {baichuan}. 
The success of LLMs has also promoted the research interest in large multimodal models (LMMs), which aim to empower LLMs with the ability to handle multiple modalities. 

\paragraph{Large multimodal models (LMMs).} The research of LMMs can be roughly categorized into two lines: loosely coupled and tightly coupled approaches. Loosely coupled approaches, e.g., Visual ChatGPT \cite {visual-chatgpt}, MM-REACT \cite {mmreact}, and HuggingGPT \cite {hugginggpt}, leverage the LLMs to coordinate with multiple external vision models to understand and express visual information. These methods are also known as {multimodal agent} approaches since the LLMs can autonomously plan and invoke vision models as tools to tackle the multimodal tasks. The tightly coupled approaches aim to train end-to-end multimodal models by aligning pretrained vision models with LLMs through multimodal interaction modules. Flamingo introduces a gated cross-attention mechanism to align the visual and language modalities \cite{flamingo}. LLaVA uses a simple linear projection layer to map the visual representations into the embedding space of the LLM \cite{llava1}. BLIP-2 \cite{blip2} and MiniGPT-4 \cite{minigpt4} learn a heavier Q-Former modules to establish alignment across modalities effectively. Subsequent studies construct diverse and high-quality multimodal instruction datasets to enhance specific capabilities of the learned LMMs \cite{instructblip, qwen-vl, shikra, llava1.5, vila, nextchat, cogvlm}. 

\paragraph{Lightweight LLMs and LMMs.} Currently, most popular LLMs and LMMs are parameter-heavy and computation-intensive, which hinders their applicability in resource-constrained scenarios like PC and mobile devices. It is worth noting that there has been an increasing interest in developing {lightweight} yet powerful LLMs and LMMs below the 7B scale. For example, Phi (1.3B and 2.7B) \cite{phi1, phi2}, Gemma (2B) \cite{gemma}, Qwen (1.8B and 4B) \cite{qwen}, TinyLlama (1.1B) \cite{tinyllama}, MobileLLaMA (1.4B and 2.7B) \cite{mobilevlm} and MiniCPM (2B) \cite{minicpm} are representatives LLMs of this size. The open-source of these lightweight LLMs also facilitates the exploration of lightweight LMMs, thus a series of early attempts have been made successively, such as TinyGPT-V (3B) \cite{tinygptv}, LLaVA-Phi (3B) \cite{llavaphi}, Vary-Toy (1.8B) \cite{varytoy}, MiniCPM-V (3B), Bunny (3B) \cite{bunny} and MobileVLM (3B) \cite{mobilevlm}. However, there is a lack of through study of the model architecture, training strategy, and training data to explore the potential capabilities of lightweight LMMs.

\section{Preliminaries}\label{sec:preliminary}
\begin{figure}
	\centering
	\begin{subfigure}[h]{0.63\columnwidth}
		\includegraphics[width=\linewidth]{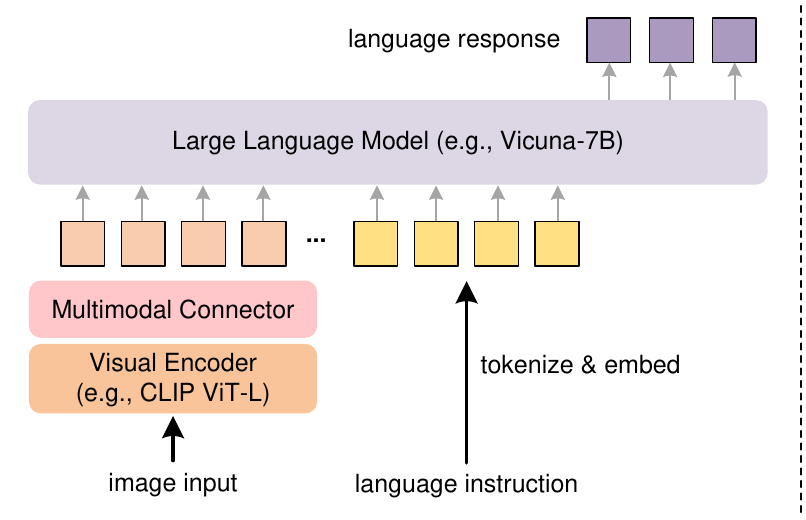}
		\vspace{-9pt}
		\captionsetup{font=footnotesize}
		\caption{\centering Model architecture}\label{fig:llava_arch}
	\end{subfigure}
	\begin{subfigure}[h]{0.32\columnwidth}
		\includegraphics[width=\linewidth]{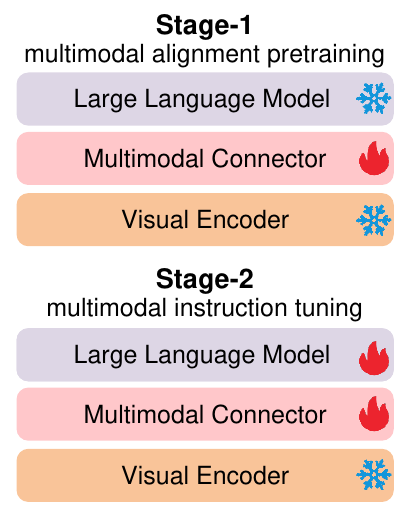}
		\captionsetup{font=footnotesize}
		\caption{\centering Training recipe}\label{fig:llava_train}
	\end{subfigure}
	\caption{LLaVA-1.5's model architecture and its two-stage training recipe.}
	\label{fig:llava}
\end{figure}

Our Imp models are derived from LLaVA-1.5 \cite{llava1.5}, a highly capable 7B LMM trained on curated publicly available data. To better describe our modifications, we first revisit LLaVA-1.5's model architecture and training recipe, as illustrated in Figure \ref{fig:llava}.

\paragraph{Model architecture.} As shown in Figure \ref{fig:llava_arch}, the architecture of LLaVA-1.5 consists of three key components: a pretrained visual encoder, a pretrained LLM, and an intermediate multimodal connector trained from scratch.
For an input image, it is represented as a sequence of visual embeddings in the same word (token) embedding space of the LLM. To achieve this goal, LLaVA introduces a visual encoder and a multimodal projector. Specifically, LLaVA-1.5 utilizes a ViT-based visual encoder pretrained by multimodal contrastive learning \cite{vit}, which refers to the CLIP ViT-L/14@336 model (0.3B) \cite{clip}, to encode the image into a sequence of 576 (24$\times$24) flattened visual features. After that, these visual features are fed through a multimodal connector module implemented by a two-layer MLP, which transforms the visual embeddings to the same dimensionality as the word embedding. The obtained visual embeddings are concatenated with the word embeddings of input language instruction to form a multimodal input, which is then fed into a pretrained LLM (e.g., Vicuna \cite{vicuna2023}) to generate a language response token by token.

\captionsetup[subfigure]{font=small}

\paragraph{Training recipe.} 
As shown in Figure \ref{fig:llava_train}, LLaVA-1.5 utilizes a two-stage training scheme, namely \textit{multimodal alignment pretraining} and \textit{multimodal instruction tuning}, to guarantee sufficient learning of each network component. Specifically, in the first stage, only the multimodal connector is trained while the rest of the model are frozen. This stage uses a collection of 558K paired image-caption data, which aims to learn the {alignment} of visual embeddings and word embeddings. In the second stage, the LLM and multimodal connector are optimized jointly while the visual encoder is still frozen. This stage aims to endow the LMM with the {instruction-following} ability by training on 665K image-instruction-response triplets collected from a mixture of academic VQA datasets and GPT-generated datasets.

\section{A Roadmap from LLaVA to Imp}

\begin{wrapfigure}{r}{0.52\textwidth}
	\centering
	\includegraphics[width=0.52\textwidth]{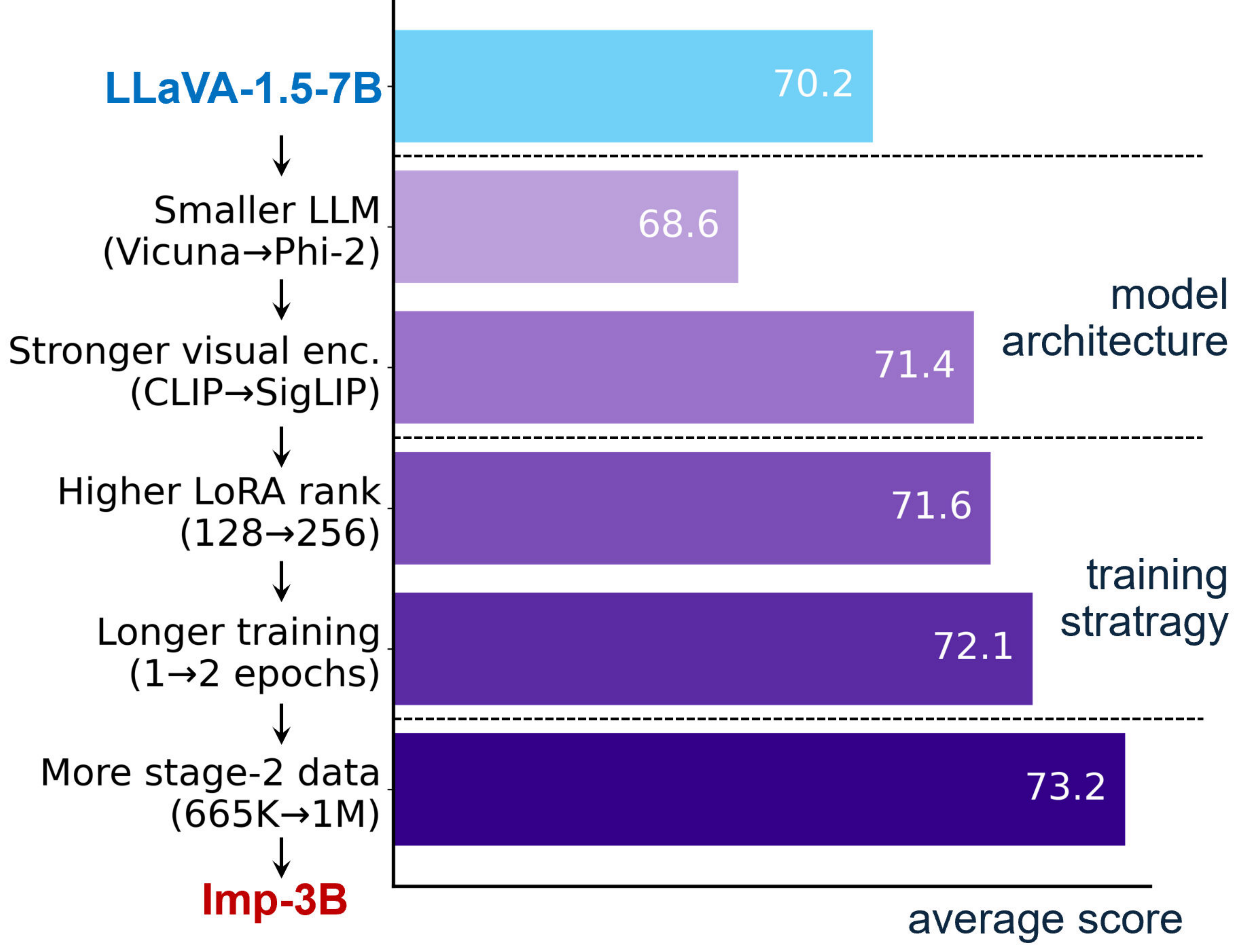}
	\vspace{-10pt}
	\caption{An overview roadmap from LLaVA-1.5-7B to Imp-3B. The average score is calculated on six commonly-used LMM benchmarks, namely VQAv2 \cite{goyal2017vqav2}, GQA \cite{hudson2019gqa}, TextVQA \cite{singh2019textvqa}, ScienceQA-IMG \cite{lu2022learn}, POPE \cite{li2023pope}, and MMB (dev) \cite{liu2023mmbench}. More detailed results can be referred to Table \ref{tab:ablations}.}
	\label{fig:roadmap}
	\vspace{-5pt}
\end{wrapfigure}

In this section, we introduce a detailed roadmap to obtain our Imp-3B model from LLaVA-1.5-7B. To remedy the capability degradation of smaller models, we conduct a thorough study of the design space of LMMs, namely model architecture, training strategy, and training data. An intuitive roadmap is shown in Figure \ref{fig:roadmap}, and its extended results are shown in Table \ref{tab:ablations}. Detailed analyses are provided below.

\subsection{Optimized Model Architectures}
We start the exploration by searching for optimal model architectures, which consists of the choices of the LLM and visual encoder. 

\paragraph{Choice of the LLM.} We adopt the LLaVA-1.5-7B model \cite{llava1.5} trained with LoRA \cite{lora} as our reference model. Based on calculations, the majority of parameters in LLaVA come from its LLM backbone Vicuna-7B \cite{vicuna2023}. In order to obtain a lightweight LMM, our first step is to replace Vicuna with a smaller yet strong LLM. To make a trade-off between efficiency and efficacy, we choose candidate LLMs at a 2.7B scale for comparison, namely, Phi-2 \cite{phi2} and MobileLLaMA \cite{mobilevlm}.

From the results in section 1.1 of Table \ref{tab:ablations}, we can see that: 1) when the same visual encoder is used  ({i.e.}, CLIP ViT-L), replacing Vicuna-7B with a smaller LLM (Phi-2 or MobileLLaMA) brings notable performance degradation in terms of average score over the six benchmarks. This suggests that the LMM's performance is highly dependent on its supporting LLM, and the capability of larger LLMs usually outperforms smaller ones according to the scaling law \cite{scalinglaw}. 2) With the same 2.7B model size, the LMM with Phi-2 significantly outperforms the counterpart with MobileLLaMA, showing the superiority of Phi-2 due to its meticulously organized training data. Therefore, we choose Phi-2 as our default LLM in the following design.

\paragraph{Choice of the visual encoder.} Apart from the LLM, the choice of the visual encoder also impacts the capability of the LMM. LLaVA-1.5 uses the powerful CLIP-ViT-L/14@336 model as its visual encoder, which is contrastively pretrained on 400M curated image-text pairs \cite{clip}. To validate the capability of different visual encoders, we experiment with two pretrained visual encoders, namely, IN1k-ViT-L/16@336 \cite{vit} and SigLIP-SO400M/14@384 \cite{siglip}. The former one is the original ViT model pretrained on ImageNet-21K and then finetuned on ImageNet-1K, and the latter one uses the same training methodology as CLIP but a shape-optimized ViT architecture with a slightly larger image resolution.

The results in section 1.2 of Table \ref{tab:ablations} show that: 1) with similar model architecture and the same input resolution, the LMM with CLIP-ViT-L significantly outperforms the counterpart with IN1k-ViT-L/16@336. This can be explained by the fact that large-scale image-text contrastive learning can facilitate generalization of the visual encoder and obtain more fine-grained visual representations. 2) Replacing CLIP-ViT with SigLIP brings consistent performance improvement on all the benchmarks, which can be explained by the synergistic effect of improved model capability and increased visual tokens (576 in CLIP \textit{vs}. 729 in SigLIP). Therefore, we choose SigLIP as our default visual encoder in the following. It is worth noting that the current 3B-scale LMM has already surpassed the reference 7B counterpart in terms of the average score, showing the feasibility and potentiality of lightweight LMMs.
\begin{table*}[h]
	\small
	\centering
	\caption{A detailed roadmap from LLaVA-1.5-7B \cite{llava1.5} to our Imp-3B model. Each section is separated by a line, which refers to one of the three aspects of the design space. The ablation studies are conducted in a {progressive} manner that each selected choice (marked with a purple background) will be used in the following experiments. The best result on each dataset is bold.}
	\label{tab:ablations}
	\scalebox{0.945}{
		\begin{tabular}{l|cccccc|c}
			\toprule
			\hiderowcolors
			& VQA$^\text{v2}$ & GQA & VQA$^\text{T}$ & SQA$^\text{I}$ & POPE & MMB & Avg. Score \\
			\hiderowcolors
			\midrule
			\textbf{1. Optimized Model Architectures}& & & & & & & \\
			\textbf{1.1. Choice of the LLM} & & & & & & & \\
			\hspace{1.5em} Vicuna-1.5 (7B) \cite{vicuna2023} (LLaVA-1.5 \cite{llava1.5}) 
			& 79.1 & 63.0 & 58.2 & 68.4 & 86.4 & 66.1 & 70.2 \\
			\hspace{1.5em} MobileLLaMA (2.7B) \cite{mobilevlm} &76.0& 60.6 & 49.9 & 58.7 & 87.0 & 58.4 & 65.1\\
			\rowcolor{mypurple}
			\hspace{1.5em} Phi-2 (2.7B) \cite{phi2}                & 76.5 & 57.2 & 52.3 & 69.8 & 87.2 & 68.6 & 68.6 \\
			\textbf{1.2. Choice of the visual encoder} & & & & & & & \\ 
			\hspace{1.5em} CLIP-ViT-L/14@336 (0.3M) \cite{clip}        & 76.5 & 57.2 & 52.3 & 69.8 & 87.2 & 68.6 & 68.6 \\
			\hspace{1.5em} IN1K-ViT-L/16@336 (0.3M) \cite{vit}        & 67.5 &55.2&38.1&63.1& 83.2 & 50.5& 59.6\\
			\rowcolor{mypurple}
			\hspace{1.5em} SigLIP-SO400M/14@384 (0.4M) \cite{siglip}    & 79.9 & 62.9 & 57.9 & 70.3 & 87.8 & 69.4 & 71.4 \\
			\midrule
			\textbf{2. Improved Training Strategies} & & & & & & & \\
			\textbf{2.1. Finetuning mechanism} & & & & & & & \\
			\hspace{1.5em} Full-parameter finetuning         & 79.3 & 61.8 & 57.1 & 70.9 & 87.4 & 70.5 & 71.2 \\
			\hspace{1.5em} LoRA finetuning, rank=128   & 79.9 & 62.9 & 57.9 & 70.3 & 87.8 & 69.4 & 71.4 \\
			\rowcolor{mypurple}
			\hspace{1.5em} LoRA finetuning, rank=256   & 79.9 & 63.0 & 57.9 & 71.0 & 87.8 & 70.0 & 71.6 \\
			\hspace{1.5em} LoRA finetuning, rank=512   & 80.0 & 62.6 & 57.4 & 71.2 & 87.7 & 70.4 & 71.5 \\
			\textbf{2.2. Number of training epochs} & & & & & & & \\ 
			\hspace{1.5em} 1 epoch                & 79.9 & 63.0 & 57.9 & 71.0 & 87.8 & 70.0 & 71.6 \\
			\rowcolor{mypurple}
			\hspace{1.5em} 2 epochs               & 81.2 & 63.8 & 59.4 & 71.2 & 87.8 & 69.3 & 72.1 \\
			\hspace{1.5em} 3 epochs               & \textbf{81.5} & {64.0} & 57.7 & 70.0 & 87.5 & 69.7 & 71.7 \\
			\midrule
			\textbf{3. Augmented Stage-2 Training Data} & & & & & & & \\ 
			\textbf{3.1. OCR \& chart-oriented data} & & & & & & & \\ 
			\hspace{1.5em} Original-665K \cite{llava1.5}                                  & 81.2 & 63.8 & 59.4 & 71.2 & 87.8 & 69.3 & 72.1 \\
			\hspace{1.5em} 643K (665K$-$22K TextCaps)  & 81.2 & \textbf{64.2} & 54.6 & 69.5 & 87.9 & 70.4 & 71.3 \\
			\rowcolor{mypurple}
			\hspace{1.5em} 675K (643K+32K OCR \& chart data)                     & 81.2 & 63.9 & 57.6 & 71.0 & 87.6 & 69.6 & 71.8 \\
			\textbf{3.2. GPT4V-annotated data} & & & & & & & \\ 
			\hspace{1.5em} 675K                      & 81.2 & 63.9 & 57.6 & 71.0 & 87.6 & 69.6 & 71.8 \\
			\hspace{1.5em} 705K (675K+30K captioning data)   & 81.2 & 64.1 & 58.5 & \textbf{73.2} & 88.0 & 70.4 & 72.6 \\
			\rowcolor{mypurple}
			\hspace{1.5em} 1M (705K+300K conversation data)   & 81.2 & 63.5 & \textbf{59.8} & {72.8} & \textbf{88.9} & \textbf{72.9} & \textbf{73.2} \\
			\bottomrule
			\hiderowcolors
		\end{tabular}
	}
\end{table*}
\subsection{Improved Training Strategies}   
After exploring the model architecture, we fix the default settings and then investigate the training strategies, including the finetuning mechanism and number of training epochs.  

\paragraph{Finetuning mechanism.} As mentioned above, LLaVA is trained using a two-stage manner. In the first stage, the visual encoder and LLM are kept frozen and only the multimodal connector is trained, while in the second stage, the LLM and multimodal connector are finetuned jointly. As the first stage only acts as an initialization, it is of less importance compared to the second stage. Therefore, we maintain the first-stage training settings in LLaVA, and explore different finetuning mechanisms in the second stage. 

Our exploration is carried out on two levels. On the macro level, we compare the traditional full-parameter finetuning and LoRA finetuning \cite{lora} mechanisms. On the micro level, we explore the LoRA finetuned models with different ranks ({i.e.}, 128, 256, and 512). From the results in section 2.1 of Table \ref{tab:ablations}, we can see that: 1) the model trained with full-parameter funetuning is inferior to the models with LoRA finetuning while requiring much more GPU memory. Therefore, we use LoRA finetuning as our training mechanism for the second stage. 2) For LoRA finetuning, increasing rank from 128 to 256 brings a 0.2 point average score improvement, while further increasing it to 512 leads to a 0.1 point decrease. Based on the above observations, we adopt LoRA finetuning with rank 256 as our finetuning mechanism in the following experiments. 

\paragraph{Number of training epochs.} LLaVA-1.5 is trained for one epoch by default. However, whether the model is sufficiently trained remains unclear. To this end, we experimented with different numbers of training epochs to verify the impact of this factor. 

As shown in section 2.2 of Table \ref{tab:ablations}, increasing training epochs from 1 to 2 brings a 0.5 point improvement in average score. This observation verifies our hypothesis that the model trained for one epoch may be undertrained. Meanwhile, further increasing training epochs from 2 to 3 leads to a 0.4 point decrease, suggesting that the 2 epochs are the optimal setting. Likewise, we set the number of training epochs to 2 in the following explorations.

\subsection{Augmented Stage-2 Training Data}	

In addition to the model architecture and training strategy, the quality and diversity of training data, especially the instruction tuning data in the second stage, plays a key role in the capabilities of LMMs \cite{deepseekvl, llava1.5}. LLaVA-1.5 elaborately constructs a 665K mixed dataset from several academic-task-oriented VQA datasets, which has been widely used as the stage-2 training data for many open-source LMMs. Inspired by \cite{llavanext}, we have considered two types of training data to augment the original 665K data, namely OCR \& chart-oriented data and GPT4V-annotated data. Detailed statistical information of the augmented dataset is illustrated in Table \ref{tab:data_mixture}.

\begin{table}
	\setlength{\tabcolsep}{3pt}
	\centering
	\caption{Statistical information of the instruction tuning dataset for Imp. Based on LLaVA's original 665K data, we remove the TextCaps dataset in line with \cite{llavanext} and then append about 32K OCR \& chart-oriented datasets and 330K GPT4V-annotated datasets, resulting in 1M mixed instruction-tuning data in total.}
	\label{tab:data_mixture}
	\scalebox{0.74}{
		\begin{tabular}{c|c|ccccc|ccc}
			\toprule
			& LLaVA-1.5 (trimmed) & \multicolumn{5}{c|}{OCR \& chart-oriented datasets (32K)} & \multicolumn{3}{c}{GPT4V-annotated datasets (330K)} \\
			\midrule
			dataset & \makecell{original-665K \cite{llava1.5}\\(w/o TextCaps \cite {textcaps})} & \makecell{DVQA\\\cite{dvqa}} & \makecell{ChartQA\\\cite{chartqa}} & \makecell{DocVQA\\\cite{docvqa}} & \makecell{AI2D\\\cite{ai2d}} & \makecell{InfographicVQA\\\cite{infographicvqa}} & \makecell{ShareGPT-4V\\\cite{sharegpt4v}} & \makecell{LAION-GPT-V\\\cite{laiongpt4v}} & \makecell{ALLaVA\\\cite{allava}} \\
			\midrule
			size & 643K & 10K & 4K & 10K & 4K & 4K & 20K & 10K & 300K \\
			\bottomrule
		\end{tabular}
	}
\end{table}

\paragraph{OCR \& chart-oriented data.} Similar to \cite{llavanext, mini-gemini}, we introduce DVQA \cite{dvqa}, ChartQA \cite{chartqa}, DocVQA \cite{docvqa}, AI2D \cite{ai2d}, and InfographicVQA \cite{infographicvqa}, which are human annotated VQA datasets that focus on reasoning about of OCR and chart in images. Meanwhile, we remove the 22K TextCaps data \cite{textcaps} from LLaVA-1.5's 665K dataset, which uses the same set of training images as TextVQA \cite{singh2019textvqa}. This allows us to better evaluate the zero-shot performance on the TextVQA benchmark. As a result, we obtain an augmented dataset with 675K samples (665K+32K$-$22K).

From the results in section 3.1 of Table \ref{tab:ablations}, we can see that: 1) removing TextCaps leads to a 4.8 point significant decline on TextVQA and a 0.8 point performance drop in the average score. This reflects the actual zero-shot performance on TextVQA. 2) Introducing OCR \& chart-oriented data brings notable improvements to TextVQA and ScienceQA, both of which require understanding and reasoning about the texts in the image. 

\paragraph{GPT4V-annotated data.}
Apart from the academic VQA datasets, high-quality instruction-tuning data about images is valuable yet often inaccessible. As an alternative, one can harness the state-of-the-art LMM (e.g., GPT-4V) to annotate responses according to input images and predefined prompts. We utilize three typical GPT4V-annotated datasets, namely ShareGPT-4V \cite{sharegpt4v}, LAION-GPT-V \cite{laiongpt4v}, and ALLaVA \cite{allava}. The first two are visual captioning datasets (30K in total), and the last one is a conversation dataset for general-purpose multimodal tasks. Note that the original ALLaVA contains 708K samples. To avoid its dominance in our mixed dataset, we randomly sample a 300K subset from ALLaVA. Finally, we obtain about an augmented dataset with 1M samples (675K+30K+300K) in total.

From the results in section 3.2 of Table \ref{tab:ablations}, we observe that both the captioning and conversation data facilitate the model capability. Their synergistic effect enables us to obtain a highly capable LMM with a 73.2 average score, which is 3 points higher than LLaVA-1.5-7B on average. We name this model {Imp-3B} and will compare it with the state-of-the-art LMMs next.

\section{Main Results}
The roadmap above ends with a combination of design choices that is generic and transferrable. In addition to Phi-2, we also apply these design choices to different lightweight LLMs, namely, Qwen-1.5 (1.8B) \cite{qwen} and Phi-3 (3.8B) \cite{phi3}, to obtain a family of lightweight LMMs Imp-2B/3B/4B\footnote{We have previously released an Imp model named \texttt{Imp-v1-3B} on HuggingFace (\url{https://huggingface.co/MILVLG/imp-v1-3b}). To avoid ambiguity, the Imp models obtained in this paper are termed \texttt{Imp-v1.5-2B/3B/4B} when released on HuggingFace. To be more precise, Imp-v1 and Imp-v1.5 use slightly different image preprocess strategies (``\textit{resize-to-square}'' for Imp-v1 and ``\textit{resize-then-pad}'' for Imp-v1.5) and different training data (665K for Imp-v1 and 1M for Imp-v1.5). }. We conduct comprehensive quantitative and qualitative comparisons with state-of-the-art LMMs to validate the efficacy and efficiency of our Imp models. Each Imp model is trained on a server with 8 A100 GPUs (40GB) and finished in less than 32 hours. 
\subsection{Quantitative Comparisons with the State-of-the-art LMMs}

\begin{table*}
	\caption{
		Comparison of our Imp models (with purple backgrounds) and open-sourced state-of-the-art LMMs on ten commonly-used VQA and LMM benchmarks. \#PT and \#FT denote the number of pretrainng and instruction-tuning samples, respectively. Benchmark names are abbreviated due to space limits: 
		VQA-v2 \cite {goyal2017vqav2}; GQA \cite {hudson2019gqa}; VisWiz \cite {gurari2018vizwiz}; SQA$^\text{I}$: ScienceQA-IMG \cite {lu2022learn}; VQA$^\text{T}$: TextVQA \cite {singh2019textvqa}; POPE \cite {li2023pope}; MME$^\text{P}$: MME-Perception \cite {fu2023mme}; MMB: MMBench (dev)\cite{liu2023mmbench}; MMB$^\text{CN}$: MMBench-Chinese (dev)\cite{liu2023mmbench}; MM-Vet \cite {yu2023mmvet}. $^*$: The training images of the datasets are observed during training. The compared LMMs are first categorized into the normal-sized and lightweight groups using a cut-off of 6B. Among the lightweight LMMs, we further categorize them into three groups based on their model sizes. Within each group of the lightweight LMMs, the best result on each benchmark is \textbf{bold}.
	}
	\setlength{\tabcolsep}{2pt}
	\centering
	\scalebox{0.65}{
		\begin{tabular}{lllll | ccccc | ccccc }
			\toprule
			Method   & LLM & Visual Enc@Res. & \#PT & \#FT & VQA$^\text{v2}$ & GQA & Vizwiz & SQA$^\text{I}$ & VQA$^\text{T}$ & POPE & MME$^\text{P}$ & MMB & MMB$^\text{CN}$ & MM-Vet \\
			\midrule
			\multicolumn{2}{l}{\textbf{Normal-sized LMM ($>$6B)}} \\
			Yi-VL-6B & Yi-6B & CLIP-H@224 & 100M & 26M &-&-&-&-&-&-&-&68.4&68.6&-\\
			mPLUG-Owl2-8B & LLaMA2-7B & CLIP-L@448 & 400M & 1.2M & 79.4 & 56.1 & 54.5 & 68.7 & 54.3 & - & 1450.2 & 64.5 & -  & 36.2 \\
			SPHINX-X-8B & InternLM2-7B &CLIP\&Dino@224  & 0 & 15M & 75.5* & 56.2* & 49.6 &  70.4 & 58.1 & 86.9 & 1260.4 & 57.9& - & -\\
			InstructBLIP-8B &Vicuna-7B &ViT-G/14@224  & 129M & 1.2M & - & 49.2* & 34.5 &  60.5 & 50.1 & - & - & - & - & -\\
			LLaVA-1.5-7B & Vicuna-1.5-7B  & CLIP-L@336 & 558K & 665K & 79.1$^*$& 63.0$^*$& 50.0 & 68.4& 58.2$^*$ & 86.4& 1476.9& 66.1& 58.9& 30.2\\ 
			LLaVA-1.5-13B & Vicuna-1.5-13B & CLIP-L@336 & 558K & 665K & {80.0}$^*$ & 63.3$^*$ & {53.6} & {71.2}& {60.2}$^*$ & 86.7& {1541.7}& {68.5}& {61.5}& {38.3}\\
			\midrule
			\hline 
			\multicolumn{2}{l}{\textbf{Lightweight LMM ($<$6B)}} \\
			Bunny-4B &Phi-3-3.8B & SigLIP-SO@384&2M &695K & \textbf{81.5*} & \textbf{63.5*} & - & 75.1 &-& 86.7 & 1495.2 & \textbf{73.5} & - &- \\
			\rowcolor{mypurple}
			\textbf{Imp-4B} & Phi-3-3.8B     & SigLIP-SO@384  & 558K & 1M & \textbf{81.5*} & \textbf{63.5*} & \textbf{51.2} & \textbf{78.3} & \textbf{60.2} & \textbf{86.9} & \textbf{1507.7} &  {73.3}	& \textbf{61.1} & \textbf{44.6}\\ 
			\midrule
			TinyGPT-V-3B & Phi-2-2.7B & E-CLIP-G@448 & 23M & 1M & - & 38.9$^*$ & 37.8 & - & - & - & - & - & - & -  \\
			LLaVA-Phi-3B & Phi-2-2.7B & CLIP-L@336 & 558K & 665K & 71.4$^*$ & -& 35.9 & 68.4 & 48.6$^*$ & 85.0 & 1335.1 & 59.8 & - & 28.9 \\
			MobileVLM-3B~$\text{}$~$\text{}$ & MobileLLaMA-2.7B & CLIP-L@336 & 558K & 665K & - & 59.0$^*$ & - & 61.0 & 47.5$^*$ & 84.9 & 1288.9 & 59.6 & - & - \\
			MiniCPM-V-3B & MiniCPM-SFT-2B & SigLIP-SO@384 & 300M & 8M & - & - & - & - & - & - & 1452 .0 & 67.9 & \textbf{65.3} & - \\
			Bunny-3B     & Phi-2-2.7B   & SigLIP-SO@384 & 2M & 695K & 79.8$^*$ & 62.5$^*$ & - & {70.9} & - & 86.8 & \textbf{1488.8} & {68.6} & - & - \\
			\rowcolor{mypurple}
			\textbf{Imp-3B} & Phi-2-2.7B   & SigLIP-SO@384  & 558K & 1M & \textbf{81.2*} & \textbf{63.5*} & \textbf{54.1} & \textbf{72.8} & \textbf{59.8} & \textbf{88.0} & 1446.4 & \textbf{72.9} & 46.7 & \textbf{43.3}\\
			\midrule
			Bunny-2B &Qwen1.5-1.8B   &  SigLIP-SO@384 &2M &695K  & 76.6*& 59.6* & - & 64.6 & - & 85.8 & 1300.8 & 59.1 & 58.5 & -\\
			Mini-Gemini-2B & Gemma-2B & CLIP-L@336 & 1.2M &1.5M & - & - & - & - & \textbf{56.2}&-& \textbf{1341} &59.8 &-& 31.1\\
			\rowcolor{mypurple}
			\textbf{Imp-2B} & Qwen-1.5-1.8B & SigLIP-SO@384  & 558K & 1M & \textbf{79.2*} & \textbf{61.9*} & \textbf{39.6} & \textbf{66.1} & 54.5 & \textbf{86.7}	& 1304.8 & \textbf{63.8} & \textbf{61.3} & \textbf{33.5}\\
			\bottomrule
		\end{tabular}
	}
	\label{tab:sota_comparison}
\end{table*}

In Table \ref{tab:sota_comparison}, we compare our Imp models to the state-of-the-art lightweight LMMs, namely TinyGPT-V-3B \cite{tinygptv}, LLaVA-Phi-3B \cite{llavaphi}, MobileVLM-3B \cite{mobilevlm}, MiniCPM-V-3B \cite{minicpm}, Bunny-3B \cite{bunny}, and Mini-Gemini-2B \cite{mini-gemini}. Moreover, we also add comparisons to the state-of-the-art normal-sized LMMs, namely LLaVA-1.5-7B/13B \cite{llava1.5}, InstructBLIP-8B \cite{instructblip}, mPlug-Owl2-8B \cite{mplugowl2}, SPHINX-X-8B \cite{sphinx}, to further demonstrate the superior capability of our Imp models.

From the results, we can see that Imp models steadily outperform all the counterparts at the same model size and is even on par with LLaVA-1.5-13B, reflecting the effectiveness of our elaborated model design. Compared with the strong competitor like MiniCPM-V-3B and Mini-Gemini-2B, Imp uses much less training data. This suggests that for lightweight LMMs, the quality of training data is more important than the quantity. Moreover, although Imp-2B exhibits consistently worse performance than the rest two Imp models, its performance on the Chinese MMB$^\text{CN}$ benchmark is prominent. As our training data is purely in English, this phenomenon implies that the bilingual understanding capability of the LLM ({i.e.}, Qwen-1.5) is successfully inherited by the LMM. 

\subsection{Quantitative Comparisons on Mobile Devices}

\begin{table*}
	\small
	\centering
	\caption{Latency and performance comparisons of MobileVLM-3B@336 \cite{mobilevlm}, Imp-3B@384 and Imp-3B@196 on different hardware platforms and quantization precision. All models are evaluated using \texttt{llama.cpp} framework with 16/8/4-bit quantilization precision. SD denotes the Snapdragon mobile chip from Qualcomm. $T_\text{VE}$ indicates the visual encoding time. $S_\text{prompt}$ and $S_\text{gen}$ measure the speed (tokens/s) of the prompt encoding and response generation stages, repetitively. $T_{total}$ refers to the entire latency to infer one sample.}
	\label{tab:overheads}
	\setlength{\tabcolsep}{3.5pt}
	\scalebox{0.95}{
		\begin{tabular}{llcc|rrrr|cc}
			\toprule
			Model@resolution & Hardware & \makecell[c]{Quant. \\ Precision} & ~\makecell{Size\\(GB)}~& \makecell{~$T_\text{VE}$~ \\ (s)} & \makecell{$S_\text{prompt}$\\(tokens/s)} & \makecell{$S_\text{gen}$\\(tokens/s)} & \makecell{$T_\text{total}$~\text{}\\(s)} & ~SQA$^\text{I}$ & MMB  \\
			\midrule
			\multirow{7}{*}{\makecell{MobileVLM-3B@336~~~~~}}
			& RTX 3090 & 16-bit & 5.6 & 0.027 & 5919.84 & 102.57 & 0.70 & 60.19 & 56.70 \\
			\cmidrule{2-10}
			& RTX 3090 & \multirow{3}{*}{8-bit} &\multirow{3}{*}{3.3}& 0.027 & 3937.64 & 137.26 & 0.57 & \multirow{1}{*}{\makecell[c]{60.19}} & \multirow{1}{*}{\makecell[c]{56.63}}\\
			& SD 8Gen3 & & & 4.93 & 17.19 & 13.79 & 21.10 \\
			& SD 888   & & & 9.70 & 11.18 & 6.85  & 37.09 \\
			\cmidrule{2-10}
			& RTX 3090 & \multirow{3}{*}{4-bit} &\multirow{3}{*}{2.1}& 0.027 & 4006.48 & 176.17 & 0.47 & \multirow{1}{*}{\makecell[c]{58.70}} & \multirow{1}{*}{\makecell[c]{55.76}}\\
			& SD 8Gen3 & & & 4.93 & 15.73 & 13.07 & 22.23 \\
			& SD 888   & & & 9.63 & 10.73 & 8.15  & 35.85 \\
			\midrule
			\multirow{7}{*}{\makecell[c]{Imp-3B@384}}
			& RTX 3090 & 16-bit & 6.0 & 0.045 & 6125.18 & 97.91 & 0.83 & 73.13 & 71.05 \\
			\cmidrule{2-10}
			& RTX 3090  & \multirow{3}{*}{8-bit} &\multirow{3}{*}{3.6} & 0.043 & 4731.30 & 138.17 & 0.68 & \multirow{1}{*}{\makecell[c]{73.03}} & \multirow{1}{*}{\makecell[c]{71.05}}\\
			& SD 8Gen3 & & & 8.57  & 15.73 & 11.91 & 63.57  \\
			& SD 888   & & & 17.52 & 10.10 & 5.82  & 105.80 \\
			\cmidrule{2-10}
			& RTX 3090  & \multirow{3}{*}{4-bit} &\multirow{3}{*}{2.3} & 0.043 & 4814.34 & 167.17 & 0.59 & \multirow{1}{*}{\makecell{72.88}} & \multirow{1}{*}{\makecell{70.88}}\\
			& SD 8Gen3 & & & 8.88  & 14.02 & 11.02 & 70.12  \\
			& SD 888   & & & 16.9  & 9.89  & 6.74  & 105.05 \\
			\midrule
			\multirow{7}{*}{\makecell{Imp-3B@196}}
			& RTX 3090  & 16-bit & 6.0 & 0.018  & 6325.2 & 101.71 & 0.69 & 68.37 & 63.40\\
			\cmidrule{2-10}
			& RTX 3090 & \multirow{3}{*}{8-bit} &\multirow{3}{*}{3.6} & 0.017 & 3744.14 & 137.57 & 0.57 & \multirow{1}{*}{\makecell[c]{68.42}} & \multirow{1}{*}{\makecell[c]{63.66}}\\
			& SD 8Gen3 & & & 2.09 & 17.12 & 13.63 & 21.24 \\
			& SD 888   & & & 4.02 & 11.68 & 6.86  & 34.65 \\
			\cmidrule{2-10}
			& RTX 3090 & \multirow{3}{*}{4-bit} &\multirow{3}{*}{2.3} & 0.017 & 3930.54 & 170.42 & 0.48 & \multirow{1}{*}{\makecell[c]{68.47}} & \multirow{1}{*}{\makecell[c]{63.14}}\\
			& SD 8Gen3 & & & 2.10 & 15.64 & 12.73 & 22.72 \\
			& SD 888   & & & 4.02 & 10.84 & 8.31  & 34.37 \\
			\bottomrule
		\end{tabular}
	}
\end{table*}

The lightweight nature of the Imp models enables us to deploy them on mobile devices. We choose two mobile phones equipped with Snapdragon 8Gen3 and Snapdragon 888 chips, respectively. To measure the latency on different devices with different quantization precision (16/8/4-bit), we resort to the open-source \texttt{llama.cpp} inference framework\footnote{\url{https://github.com/ggerganov/llama.cpp}}. 

The inference process of these LMMs consists of three key stages: visual encoding, prompt encoding, and response generation. The total inference time can be calculated as follows: 
\begin{equation}
	T_\text{total} = T_\text{VE} + S_\text{prompt}/N_\text{prompt} + S_\text{gen}/N_\text{gen} + T_\text{other}
\end{equation}
where $T_\text{total}$ and $T_\text{VE}$ refer to the entire time and visual encoding time consumed by a single inference. $S_\text{prompt}$ and $S_\text{gen}$ measure the speed (tokens/s) of the prompt encoding and response generation stages, repetitively. $N_\text{prompt}$ and $N_\text{gen}$ refer to the number of prompt tokens (text tokens plus image embeddings) and generated tokens, respectively. $T_\text{other}$ means the rest negligible overheads. Note that the model loading time is not counted in the total time.

We compare the latency and performance of MobileVLM-3B@336 and Imp-3B@384 in Table \ref{tab:overheads}. To further accelerate our Imp-3B model, we retrain a variant Imp-3B@196 with a reduced image resolution of 196$\times$196 pixels. For latency measurement, all models are fed with the same text prompt (41 tokens) and generate the response with the same length (stops at the 64th token). For performance evaluation on SQA$^\text{I}$ \cite{lu2022learn} and MMB \cite{liu2023mmbench}, the results of models with different quantization precision are computed using a RTX 3090 GPU for efficiency\footnote{Tolerable numerical differences are observed in the evaluated results obtained by \texttt{llama.cpp} and \texttt{PyTorch} frameworks.}. Note that the number of image embeddings is different for the three compared models, which are 576, 729, and 196 for MobileVLM-3B@336, Imp-3B@384, and Imp-3B@196, respectively. From the results, we have the following observations.

First, the inference latency of models on GPUs and mobile devices differs significantly in magnitude, which is about 100$\times$ difference in visual encoding and prompt encoding, and about 10$\times$ difference in response generation. Similarly, there exists a 2$\sim$3$\times$ difference in inference speed between two mobile devices. These divergences reveal the challenges for the deployment of LMM at the edge.

Second, low-bit quantization techniques can effectively reduce model size, making it easier to be stored and run on mobile devices. Moreover, compared with the 16-bit unquantized version, 4-bit or 8-bit quantization lead to subtle performance and latency degradation.

Third, our Imp-3B@384 is slower than MobileVLM-3B@336 at each stage, which is mainly caused by more visual embeddings (729 \textit{vs}. 576). To address this issue, we simply reduce the input image resolution and obtain an Imp-3B@196 variant. Imp-3B@196 achieves a comparable speed to MobileVLM-3B@336, while performing better in terms of accuracies on ScienceQA-IMG \cite{lu2022learn} and MMB \cite{liu2023mmbench}.

To summarize, Imp-3B@196 with 4-bit quantization achieves a good balance in model storage, latency, and capability. We regard it as our default model for the following deployment.  

\subsection{Qualitative Results}

To demonstrate the comprehensive capabilities of our Imp models and understand how they behave in real-world scenarios, we provide some visualized examples by comparing the response of our Imp models with the larger LLaVA-v1.5-7B model. 

\begin{figure}[htbp]
	\centering
	\vtop{\includegraphics[width=0.97\linewidth]{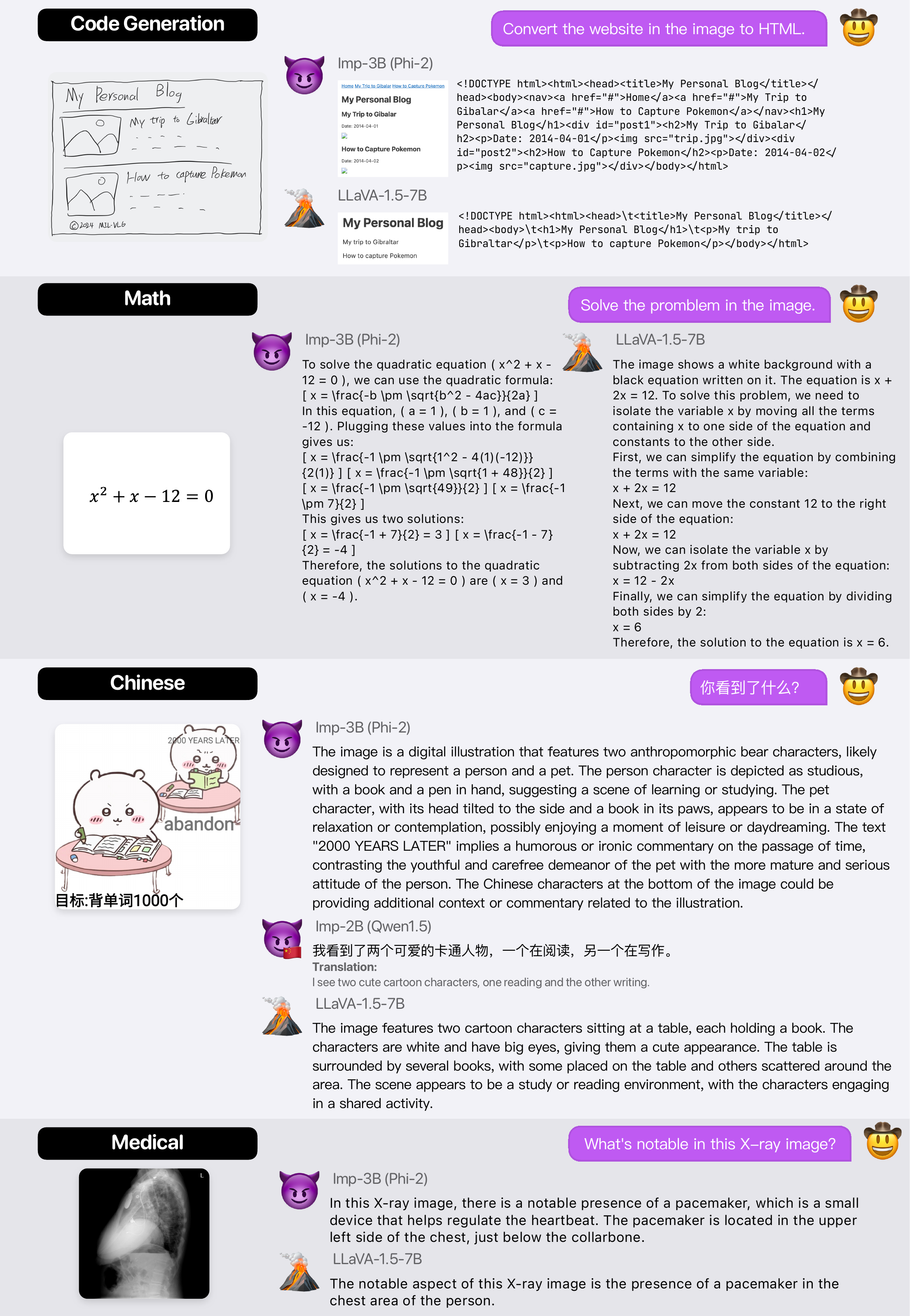}}
	\caption{Comprehensive skill demonstrations of Imp, including code generation, math problem solving, Chinese conversation, and medical image understanding.}
	\label{fig:example1}
\end{figure}

\begin{figure}[htbp]
	\centering
	\vtop{\includegraphics[width=0.93\linewidth]{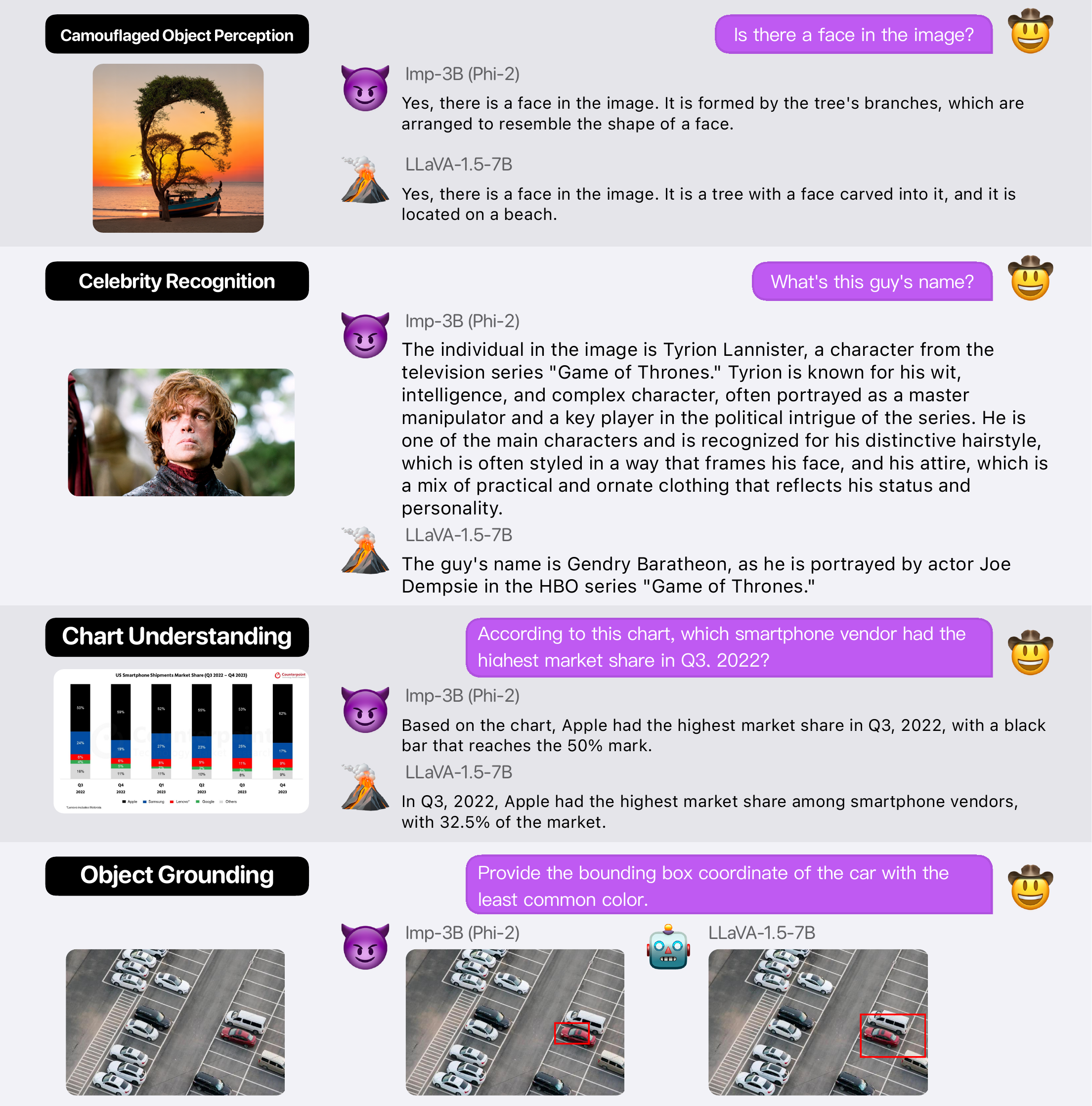}}
	\vspace{-5pt}
	\caption{More diverse skill demonstrations of Imp, including camouflage perception, celebrity recognition, chart understanding, and object grounding.}
	\label{fig:example2}
\end{figure}

\begin{figure}[htbp]
	\centering
	\vtop{\includegraphics[width=0.93\linewidth]{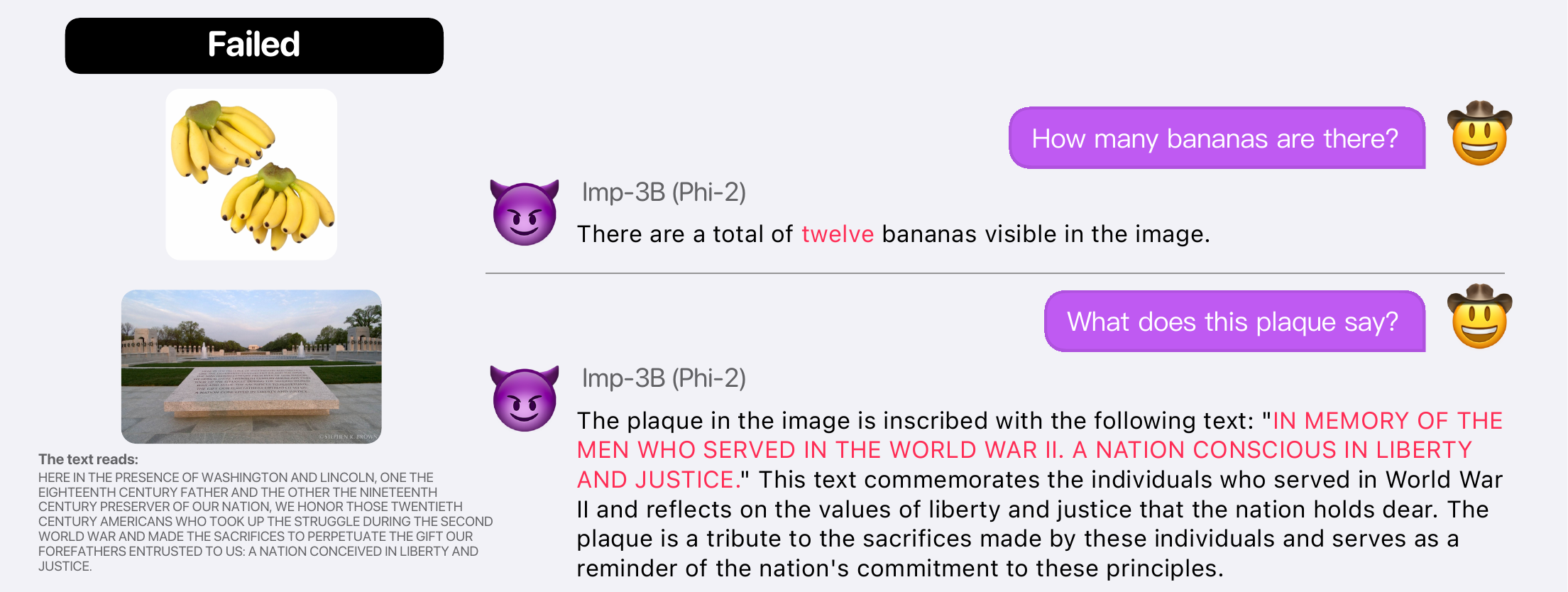}}
	\vspace{-5pt}
	\caption{Failure cases of Imp, including the hard examples about dense object counting, and long-text OCR.}
	\label{fig:fail-example}
\end{figure}

Figure \ref{fig:example1} demonstrates different skills of Imp models, including code generation, math problem solving, Chinese conversation, and medical image understanding. While the LLaVA-v1.5-7B model generates unsatisfactory outputs in these cases, the Imp-3B model provides more reasonable responses, showing its superiority in VL understanding and reasoning, as well as the completeness of knowledge. For Chinese conversation, only the Imp-2B (Qwen-1.5) responds in the expected language, while the other two models fail to generate Chinese responses. This indicates the importance of the multilingual LLMs when building more user-friendly LMMs. For medical image understanding, although Imp-3B has not been deliberately trained on medical images, it exhibits impressive zero-shot capabilities.

Figure \ref{fig:example2} shows more diverse skill demonstrations of our Imp models, including camouflage perception, celebrity recognition, chart understanding, and object grounding. In these cases, Imp-3B can consistently provide correct and coherent responses, while LLaVA-v1.5-7B gets responses of relatively poorer quality. Those comparisons demonstrate the strength of Imp models in solving diverse tasks and assisting users with various intentions.

Finally, we showcase some failure cases of Imp models in Figure \ref{fig:fail-example}, including the hard examples about dense object counting, and long-text OCR. These failure cases demonstrate the weakness of Imp models and point out the directions for future improvement.

\section{ImpChat: LMM Assistants on Multiple Platforms}

\captionsetup[subfigure]{font=small}
\begin{figure}
	\centering
	\begin{subfigure}[h]{0.77\columnwidth}
		\includegraphics[width=\linewidth]{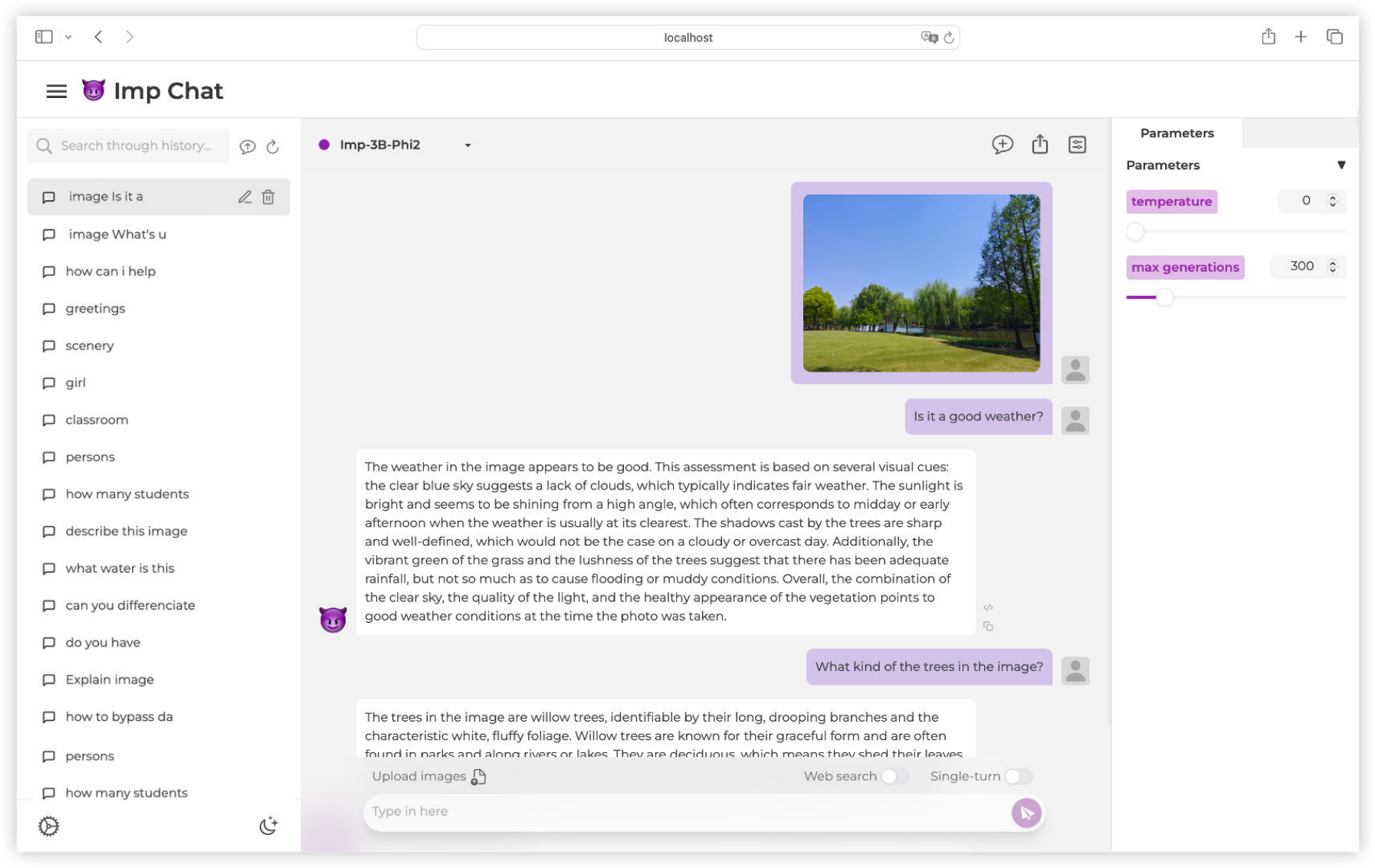}
		\vspace{-9pt}
		\captionsetup{font=footnotesize}
		\caption{\centering ImpChat-Web}\label{fig:app1}
	\end{subfigure}
	\begin{subfigure}[h]{0.22\columnwidth}
		\includegraphics[width=\linewidth]{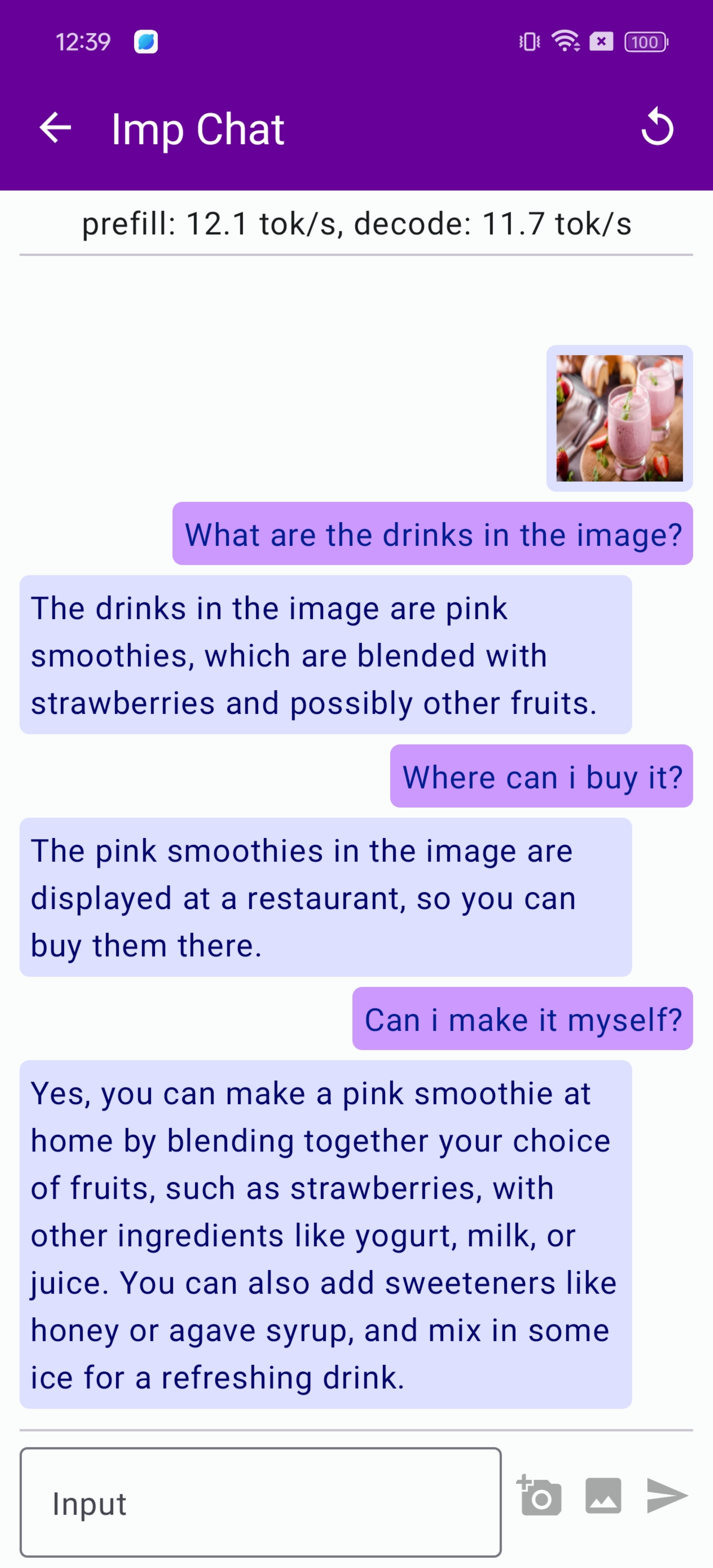}
		\captionsetup{font=footnotesize}
		\caption{\centering ImpChat-Mobile}\label{fig:app2}
	\end{subfigure}
	\caption{The UI demonstrations of ImpChat on multiple platforms.}
	\label{fig:app}
\end{figure}

Based on the Imp models we trained and the deployment strategies we explored, we have developed a software suite of conversational AI interfaces named {ImpChat}, which is dedicated to providing users with an easy-to-use LMM assistant in {offline} environments. ImpChat is capable of understanding text and images in a wide range of conversational contexts, and generating coherent responses that are engaging and informative. As shown in Figure \ref{fig:app}, ImpChat provides specialized interfaces for different platforms, {i.e.}, web and mobile devices.

For web users, we offer ImpChat-Web, a web-based chat assistant. The frontend of ImpChat-Web is implemented by ChuanhuChat\footnote{\url{https://github.com/GaiZhenbiao/ChuanhuChatGPT}}, a popular UI framework for open-source LLM deployment, which is also developed by our team. The backend is served by a server with multiple GPUs, with an optimized load balancing strategy to offer a high throughput and low latency for concurrent user accesses. 

For mobile phone users, we develop ImpChat-Mobile based on the MLC framework\footnote{\url{https://github.com/mlc-ai/mlc-llm}}. ImpChat-Mobile is an application that can be installed across different platforms such as Android and iOS. ImpChat-Mobile is run in an offline environment without an Internet connection, which is more favorable for data privacy scenarios. We currently implement ImpChat-Mobile for Android only.

\subsection{Conclusion and Future Work}

In this paper, we present a family of lightweight LMMs named Imp. To balance the model efficiency and efficacy, we conduct a systematic exploration and establish a comprehensive roadmap from the aspects of model architecture, training strategy, and training data. Based on the empirical studies and different lightweight LLMs, we obtained Imp models at the 2B$\sim$4B scales. Notably, our Imp-3B model steadily outperforms all the existing lightweight LMMs of similar size and even surpasses the state-of-the-art LMMs at the 13B scale. Our optimized Imp models can run on a mobile phone with a high inference speed. 

In the future, we consider improving the model capability from following aspects: 1) appending high-quality training data for specific tasks, such as Chinese OCR and set-of-marks instruct-tuning data, 2) introducing more effective training strategy such as knowledge distillation and human-aligned preference optimization, 3) exploring more efficient model compression algorithms such as 1-bit and 1.58-bit quantization methods, and 4) supporting broader input and output modalities such as audio and 3D. On the application level, we will explore the usage Imp models in more diverse scenarios such as UI agent and robot manipulation. We hope our imp project could establish an ecosystem to facilitate future research and applications with lightweight LMMs.

\section*{Acknowledgements}
This work was supported in part by the National Natural Science Foundation of China under Grants No.62125201, No.62072147, No.62020106007, No.62206082, and in part by the Zhejiang Provincial Natural Science Foundation of China under Grants No.LR22F020001 and No.DT23F020007.

%

\bibliographystyle{abbrvnat}
\bibliography{imp}

\end{CJK*}
\end{document}